\documentclass{article}

\usepackage{arxiv}

\usepackage[utf8]{inputenc} 
\usepackage[T1]{fontenc}    
\usepackage{hyperref}       
\usepackage{url}            
\usepackage{booktabs}       
\usepackage{amsfonts}       
\usepackage{nicefrac}       
\usepackage{microtype}      
\usepackage{graphicx}
\usepackage{enumitem}
\usepackage{tabularx}
\usepackage[numbers]{natbib}
\usepackage{hhline}

\newcommand\clearrow{\global\let\rowmac\relax}
\newcommand{\cls}[1]{\texttt{#1}}
\newcolumntype{Y}{>{\centering\arraybackslash}X}
\clearrow

\title{CPPE - 5: Medical Personal Protective Equipment Dataset}

\author{
 Rishit Dagli \\
 Department of Computer Science\\
 University of Toronto\\
 40 St. George Street, Toronto, ON M5S 3H4, Canada\\
  \texttt{rishit.dagli@mail.utoronto.ca} \\
   \And
 Ali Mustufa Shaikh \\
 Senior Developer Advocate\\
  Postman Inc.\\
   309, Venkatesh complex, 560038, India\\
  \texttt{ali.shaikh@postman.com} \\
}

\begin{document}
\maketitle
\begin{abstract}
We present a new challenging dataset, CPPE - 5 (Medical Personal Protective Equipment), with the goal to allow the study of subordinate categorization of medical personal protective equipments, which is not possible with other popular data sets that focus on broad-level categories (such as PASCAL VOC, ImageNet, Microsoft COCO, OpenImages, etc). To make it easy for models trained on this dataset to be used in practical scenarios in complex scenes, our dataset mainly contains images that show complex scenes with several objects in each scene in their natural context. The image collection for this dataset focuses on: obtaining as many non-iconic images as possible and making sure all the images are real-life images, unlike other existing datasets in this area. Our dataset includes 5 object categories (coveralls, face shields, gloves, masks, and goggles), and each image is annotated with a set of bounding boxes and positive labels. We present a detailed analysis of the dataset in comparison to other popular broad category datasets as well as datasets focusing on personal protective equipments, we also find that at present there exist no such publicly available datasets. Finally, we also analyze performance and compare model complexities on baseline and state-of-the-art models for bounding box results. Our code, data, and trained models are available at \url{https://git.io/cppe5-dataset}.
\end{abstract}


\section{Introduction}
\label{Introduction}

Deep learning is revolutionizing multiple areas of computer vision. An explosive popularity in this field was brought after the ImageNet Large Scale Visual Recognition Challenge (ILSVRC) \cite{russakovsky2015imagenet} and has pushed forward the state of the art in
generic object detection. It contains a detection challenge using ImageNet images \cite{deng2009imagenet}. Since then, the performance of models has been improving at unparalleled speeds. Among the many challenges in machine learning, data collection is becoming one of the critical bottlenecks \cite{roh2019survey}. As deep learning becomes popular the core of their success is the need for rich and large annotated training data \cite{Goodfellow-et-al-2016}. Larger and richer annotated datasets are a boon for leading-edge research in computer vision to enable the next generation of state-of-the-art algorithms \cite{kuznetsova2020} and have been instrumental in driving progress in object recognition over the last decade \cite{krizhevsky2012imagenet, sermanet2014overfeat, viola2001robust, redmon2016you}.

Object detection is a fundamental problem of computer vision that deals with detecting instances of visual objects of a certain class in digital images. The objective of object detection aims to develop models and techniques to provide the information: "what objects are where?" \cite{zou2019object} Building larger and richer datasets often play a key role in allowing computers to identify and interpret images as compositions of one or multiple objects which has been quite tricky for machines so far \cite{geirhos2018comparing}. Through this object detection dataset, we majorly aim to advance machines to automatically identify where objects (personal protective equipments) are precisely located.

In object detection, a number of well-known datasets and benchmarks have been released in the past 10 years. Most datasets contain a wide variety of common-level classes, such as different kinds of animals or inanimate things. Several such datasets have emerged as standards for the community including MIT-CSAIL \cite{torralba2004sharing}, PASCAL VOC Challenges (e.g., VOC2007, VOC2012) \cite{everingham2010pascal, everingham2015pascal}, ImageNet \cite{deng2009imagenet}, Caltech-256 \cite{griffin2007caltech}, Microsoft COCO \cite{lin2014microsoft} and DOTA \cite{Xia_2018_CVPR, ding2021object}. However, this dataset was built bearing in mind to allow for subordinate categorization especially for detecting personal protective equipment which is not possible with other large-scale popular datasets that focus on rather broad categories.

Though the first part subset of the dataset was released to facilitate working on Medical Personal Protective Equipments, these were carefully ported to create the final dataset expanding the goals to medical personal protective equipments. COVID-19 is causing widespread morbidity and mortality globally. The severe acute respiratory syndrome coronavirus 2 (SARS-CoV-2) responsible for this disease infected more than 17 million people by August 2020 \cite{merow2020seasonality}. It has also been observed that the global trend is approximately exponential, at a rate of 10-fold every 19 days \cite{li2021covid}. Considering this, it is very important to be able to accurately detect Medical Personal Protective Equipment to help limit the growth of COVID-19. To encourage the development of such tools we present this dataset publicly on GitHub and a subset of the dataset on Kaggle \footnote{https://www.kaggle.com/ialimustufa/object-detection-for-ppe-covid19-dataset} focusing on accurately identifying the Personal Protective Equipments through images.

In this paper, we introduce the CPPE - 5 (Medical Personal Protective Equipment), an object detection dataset, which contains images and ground-truth annotations for the task of object detection. The majority of the images have been collected from Flickr \footnote{https://www.flickr.com/}, with an aim to collect a majority of non-iconic images. A small portion of images was collected from Google Images as well. After doing so each of the images was annotated using crowd-sourcing techniques \cite{vaughan2017making}. Each of these annotations were evaluated by multiple people and were also then evaluated by us to keep a strict check on the quality of the ground truth annotations.

\begin{figure}
    \centering
    \includegraphics[width=\textwidth]{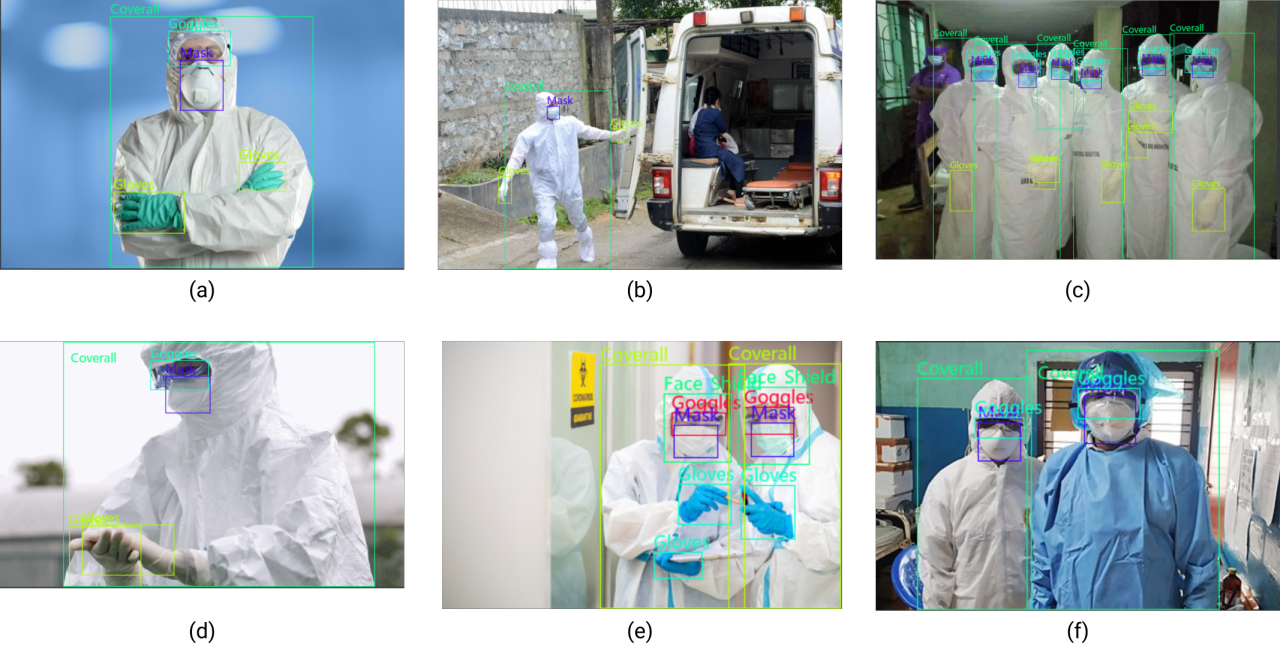}
    \caption{Example annotations in CPPE - 5 for object detection demonstrating the five classes of our data set. Each example image is shown with an outline (bounding box) and the object it is identified as.}
  \label{fig:annotation-ex}
\end{figure}

As mentioned, we provide unified annotations for the task of object detection with the dataset. In Fig. \ref{fig:annotation-ex} we show examples of annotations provided in the dataset. In (a) coveralls, gloves, mask, goggles (b) coveralls, gloves, mask (c) coveralls, gloves, goggles (d) coveralls, gloves, mask, goggles (e) coveralls, gloves, mask, goggles, face shield (f) coveralls, mask, goggles are demonstrated. Some more sample images for each category could be found in Appendix \ref{appendix:sample-images}.

With the CPPE-5 dataset, we hope to facilitate research and use in applications at multiple public places to autonomously identify if a PPE kit has been worn and also which part of the PPE kit has been worn. One of the main aims of this dataset was to also capture a higher ratio of non-iconic images or non-canonical perspectives \cite{CUTZU19943037} of the objects in this dataset. We further hope to see high use of this dataset to aid in medical scenarios which would have a huge effect worldwide.

The remainder of this article is organized as follows: In Sect. \ref{Related Work} related works are given. In Sect. \ref{Dataset Collection and Annotation} we describe the process used to collect and annotate the dataset. In Sect. \ref{Dataset Statistics} we present statistics related to the dataset. In Sect. \ref{Experimental Results} we present the experimental results, training multiple state-of-the-art and baseline models. In Sect. \ref{Conclusion} we conclude the article and give future works.

\section{Related Work}
\label{Related Work}

Throughout the history of computer vision research rich and large datasets have played a very important role. They not only provide a means to train and evaluate algorithms, but they also drive research in new and more challenging directions \cite{lin2014microsoft}. Earlier datasets like the Caltech-256 Object Category Dataset \cite{griffin2007caltech} and the MIT Pedestrian Database \cite{papageorgiou2000trainable} facilitated the direct comparison of hundreds of computer vision algorithms and also pushed toward more complex problems. Recent datasets like The Open Images dataset v4 with \(\sim 9.2\) M images \cite{kuznetsova2020} ImageNet dataset \cite{deng2009imagenet} with \(\sim 14\) M images and Microsoft COCO with \(\sim 2.5\) M labeled instances \cite{lin2014microsoft} have enabled breakthroughs in object detection research with a new wave of deep learning algorithms.

Performing object detection often requires identifying which specific class the object belongs to and also localizing the object in the image usually done with a bounding box as shown in Fig. \ref{fig:annotation-ex}. One of the earliest algorithms focused on face detection often using ad hoc datasets \cite{hjelmaas2001face}. Later more realistic and challenging datasets were built which facilitated the creation of many deep-learning algorithms. Transformers \cite{vaswani2017attention} were first introduced to vision in Vision Transformer (ViT) \cite{dosovitskiy2020image} by splitting an image into a sequence of visual tokens. The self-attention strategy in ViTs has demonstrated superior performance to modern convolutional neural networks (ConvNets) when trained with optimized recipes. A lot of popularity in using Transformers for object detection tasks was brought through DEtection TRansformer (DETR) \cite{carion2020end} and achieved at-par results with earlier methods like Faster RCNN \cite{7485869}. After this multiple works tried training transformers for object detection mainly using ViTs directly for object detection \cite{dosovitskiy2020image} and Swin Transformers \cite{liu2021swin}. Recently self-attention and transformer-based methods have shown a lot of promise for object detection and dominated the state-of-the-art for this task \cite{Liu_2022_CVPR,https://doi.org/10.48550/arxiv.2203.03605,https://doi.org/10.48550/arxiv.2205.14141,https://doi.org/10.48550/arxiv.2211.03594,https://doi.org/10.48550/arxiv.2211.12860}.

For the detection of basic object categories the PASCAL VOC datasets \cite{everingham2010pascal} were created which contained \(20\) object categories, over \(11,000\) images, and over \(27,000\) annotated objects using bounding boxes of which almost \(7,000\) had detailed segmentations. Later the ImageNet dataset was created \cite{deng2009imagenet} which included over \(14\) M images across \(1000\) object categories. The ImageNet large-scale visual recognition challenge facilitated the creation of many deep learning algorithms namely AlexNet \cite{krizhevsky2012imagenet}, Inception v1 \cite{szegedy2015going}, VGGNet \cite{simonyan2015deep}, ResNet \cite{he2016deep} and more. Later the Microsoft COCO: Common Objects In Context dataset \cite{lin2014microsoft} was created for the detection and segmentation of objects occurring in their natural context. This dataset aimed to find non-iconic images containing objects in their natural context. The COCO dataset consists of over \(330,000\) across \(91\) categories with \(1.5\) M object instances.

Machine Learning for Health is quite a popular field with quite a lot of research pertaining to Machine Learning for COVID-related topics \cite{kushwaha2020significant, alimadadi2020artificial, 10.1371/journal.pone.0235187}. Many prior works aim to solve a binary classification problem: often if a mask is worn or not; masks are one of the most widely used components of a personal protective equipment kit. The datasets acquired in these papers were in controlled environments or simulated images however to deploy these tools majorly requires them to be robust to multiple variations (eg. lighting conditions, terrain, and background objects). In the next part of this section, we talk about some related work about identifying masks in images, masks being one of the most widely used objects and are also present in our dataset. However, to the best of our knowledge, we found no related work for the rest of the categories in our dataset.

Chowdary et al. \cite{chowdary2020face} in their paper transfer learn on top of Inception V3 pre-trained on ImageNet dataset \cite{szegedy2015inceptionnv3} for the task of binary classification: identifying if a mask has been worn or not. This paper also claims to achieve quite plausible results in testing on simulated data. However, the models proposed in this paper were trained and tested on simulated data: where an image of a mask was artificially superimposed later on top of the face images. Furthermore, the images in this dataset are all iconic face images on top of which a mask was artificially added, this tends to lose out not only on a lot of contextual information but models trained on this data are unable to identify all kinds of mask and masks worn in different positions due to the artificial training data. To this end, in our dataset, we have ensured each image is a real image and no objects were artificially added on the image. Our dataset also focuses on more than one category of personal protective equipment unlike this dataset which focuses on only masks.

Wang et al. \cite{wang2020masked} in their paper introduce three datasets Masked Face Detection Dataset (MFDD), Real-world Masked Face Recognition Dataset (RMFRD), and Simulated Masked Face Recognition Dataset (SMFRD) for the task of binary classification. The multi-granularity masked face recognition model developed in this paper also claims to achieve \(95\) \% accuracy on the Real-world Masked Face Recognition dataset. The Real-world Masked Face Recognition dataset includes \(5,000\) pictures of \(525\) people wearing masks, and \(90,000\) images of the same \(525\) subjects without masks. However, the images in this dataset are not necessarily medical masks. As an example, this dataset also includes images with a scarf worn or sports helmets and masks under the category of people wearing masks. In our Medical Personal Protective Equipment (CPPE-5) dataset, as we later mention in Section \ref{Dataset Collection and Annotation}, all the images have been checked for quality and relevance.

Loey et al. \cite{LOEY2021108288} in their paper also use the three above-mentioned datasets: Face Detection Dataset (MFDD), Real-world Masked Face Recognition Dataset (RMFRD), and Simulated Masked Face Recognition Dataset (SMFRD) to train a binary classifier. The model proposed in this paper uses ResNet-50 \cite{he2016deep} as a feature extractor and then uses traditional machine learning algorithms for classification. In this setting, the paper reports quite plausible performance on the Real-world Masked Face Recognition Dataset. However, this paper measures the performance of models by training on the Real-world Masked Face Recognition Dataset but majorly tests their models on simulated mask images and not real-world mask images. Our Medical Personal Protective Equipment (CPPE-5) dataset ensures all images are real-life images.

Nath et al. \cite{nath2020deep} in their paper aim to build a system to verify the Personal Protective equipment compliance of a construction worker. They also present an in-house dataset Pictor-v3 in this paper which contains 774 annotated images collected with crowd-sourcing techniques and 698 annotated images collected through web mining. In one of their approaches in this paper where their algorithm simultaneously detects individual workers and verifies PPE compliance with a single convolutional neural network is reported to achieve \(72.3\) \% mean average precision (mAP) in real-world settings. However, their dataset only includes three categories: worker, hat, and vest out of which only two are protective equipment categories: hat and vest. These object categories are also not well suited for medical scenarios. Our Medical Personal Protective Equipment (CPPE-5) dataset contains \(5\) categories of personal protective equipment, all of which are well suited for medical purposes.

\section{Dataset Collection and Annotation}
\label{Dataset Collection and Annotation}

\begin{table}
 \caption{Categories in the CPPE-5 dataset}
  \centering
    \begin{tabularx}{\textwidth}{l|X}
    \toprule
    coveralls   & Coveralls are hospital gowns worn by medical professionals in order to provide a barrier between patient and professional, these usually cover most of the exposed skin surfaces of the professional medics. \\ \hline
    mask        & Mask prevents airborne transmission of infections between patients and/or treating personnel by blocking the movement of pathogens (primarily bacteria and viruses) shed in respiratory droplets and aerosols into and from the wearer's mouth and nose. \\ \hline
    face shield & Face shield aims to protect the wearer's entire face (or part of it) from hazards such as flying objects and road debris, chemical splashes (in laboratories or in the industry), or potentially infectious materials (in medical and laboratory environments). \\ \hline
    gloves      & Gloves are used during medical examinations and procedures to help prevent cross-contamination between caregivers and patients. \\ \hline
    goggles     & Goggles, or safety glasses, are forms of protective eyewear that usually enclose or protect the area surrounding the eye in order to prevent particulates, water or chemicals from striking the eyes. \\
    \bottomrule
  \end{tabularx}
  \label{tab:categories}
\end{table}

This section describes how we decided on the categories and collected the images in the Medical Personal Protective Equipment - 5 dataset.

\subsection{Object Categories}
\label{Object Categories}

To create a dataset we had to ensure the categories we choose from a representative set of all categories, be relevant to practical applications, and occur with high enough frequency to enable the collection of a large dataset. A small group of daily Medical Personal Protective Equipment users were asked to share components of a PPE kit based on how often they are used and their usefulness for practical applications. Through this, we received 7 potential categories for this dataset: coveralls or gowns, masks, face shields, gloves, shoe covers, respirators, and goggles

Some common PPE objects which are quite similar to the above list like lab coats, safety boots, full facepiece respirators, self-contained breathing apparatus, etc. were not included in the initial list of potential categories. Also, we omitted some PPE objects which are not used for medical scenarios from the initial categories; like helmets, harnesses, hearing protection, ballistic vests, etc. to maintain the focus of this dataset.

The final selection of categories attempts to pick categories for which obtaining a large number of images with categories in them was available. The final categories based on this did not include respirator and shoe cover due to a lack of rich annotations and enough data for these categories. The final object categories are denoted in the dataset as:

\begin{itemize}
    \item Coveralls
    \item Face\_Shield
    \item Gloves
    \item Goggles
    \item Mask
\end{itemize}

We also show in detail about the categories in this dataset in Table \ref{tab:categories}. The category definitions shown in Table \ref{tab:categories} were adapted from their Wikipedia \footnote{https://www.wikipedia.org/} pages and were also used while annotating the datasets as shown in later sections.

\subsection{Image Collection}
\label{Image Collection}

\begin{figure}
    \centering
    \includegraphics[width=\textwidth]{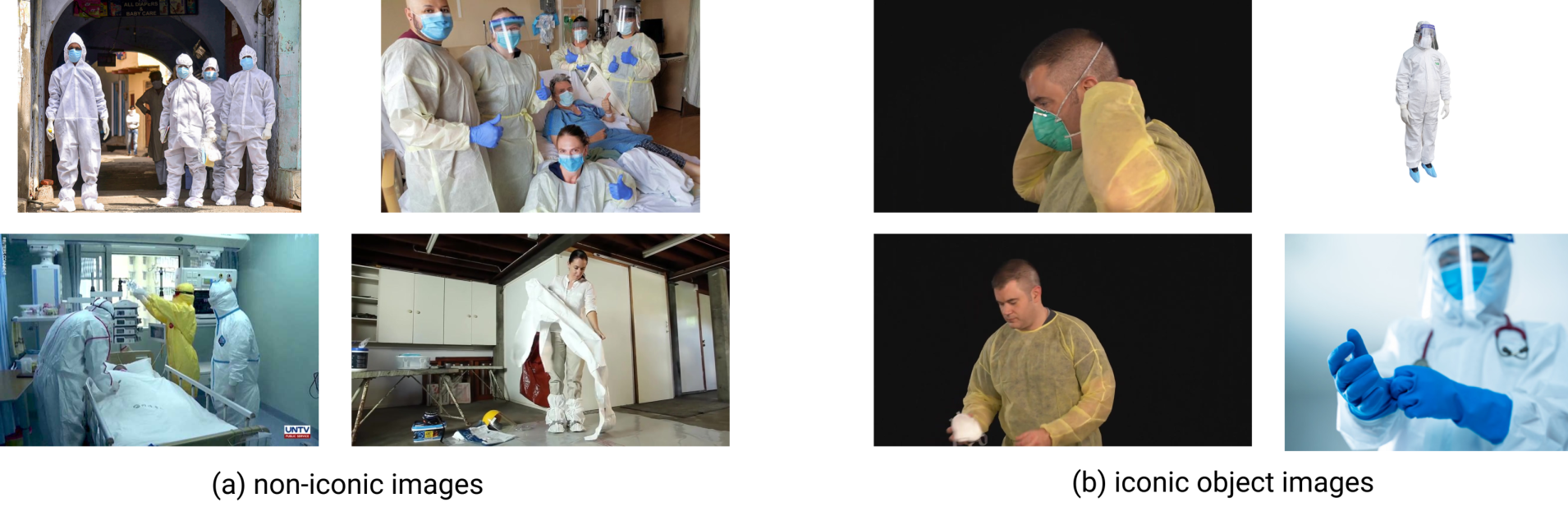}
    \caption{Example of (a) non-iconic images and (b) the little number of iconic images from our dataset.}
  \label{fig:iconic-and-non-iconic}
\end{figure}

Having decided on the object categories our next goal was to collect a set of candidate images. We classify images into two categories: iconic object images and non-iconic images as shown in Fig. \ref{fig:iconic-and-non-iconic}. While iconic images Fig. \ref{fig:iconic-and-non-iconic}(b): which have a single large object in a canonical perspective usually contain high-quality object instances they can lack important contextual information, these could be found directly by searching for the object category on Google Images \footnote{https://images.google.com/} or Bing Image Search \footnote{https://www.bing.com/images}. It has been shown by Torralba et al. \cite{5995347} that non-iconic images are better at generalizing. We thus aimed to collect a majority of non-iconic images Fig. \ref{fig:iconic-and-non-iconic}(a). This allows us to have a majority of complex images which contain several other objects.

As popularized by Caltech-UCSD Birds-200 \cite{wah2011caltech, welinder2010caltech}, Microsoft COCO \cite{lin2014microsoft} and Open Images v4 \cite{kuznetsova2020} datasets we majorly collected images from Flickr which tend to have lesser iconic images. Flickr contains images uploaded by millions of photographers with searchable metadata. A smaller portion of images was also collected from Google Images. We also remove near-duplicate images in the dataset using GIST descriptors \cite{10.1145/1646396.1646421, murillo2012localization} greatly minimizing the chances of near-duplicate images in the dataset.

The images in the CPPE-5 dataset were collected using the following process:

\begin{description}
   \item[Obtain Images from Flickr:] Following the object categories we identified earlier, we first download images from Flickr and save them at the "Original" size. On Flickr, images are served at multiple different sizes (Square 75, Small 240, Large 1024, X-Large 4K, etc.), the "Original" size is an exact copy of the image uploaded by the author. In section \ref{Dataset Statistics} we talk more about the variation in image sizes and present statistics for the sizes of images in this dataset.
   \item[Extract relevant metadata:] Flickr contains images each with searchable metadata, we extract the following relevant metadata:
    \begin{itemize}
     \item A direct link to the original image on Flickr
     \item Width and height of the image
     \item Title given to the image by the author
     \item Date and time the image was uploaded on
     \item Flickr username of the author of the image
     \item Flickr Name of the author of the image
     \item Flickr profile of the author of the image
     \item The License image is licensed under
     \item MD5 hash of the original image
    \end{itemize}
   \item[Obtain Images from Google Images:] Due to the reasons we mentioned earlier, we only collect a very small proportion of images from Google Images. For this set of images, we extract the following metadata:
    \begin{itemize}
     \item A direct link to the original image
     \item Width and height of the image
     \item MD5 hash of the original image
    \end{itemize}
   \item[Filter inappropriate images:] Though very rare in the collected images, we also remove images containing inappropriate content using the safety filters on Flickr and Google Safe Search.
   \item[Filter near-similar images:] We then remove near-duplicate images in the dataset using GIST descriptors \cite{douze2009evaluation}.
\end{description}

\subsection{Image Annotation}
\label{Image Annotation}

In this section, we describe how we annotated our image collection. The dataset was labeled in two phases: the first phase included labeling \(416\) images and the second phase included labeling \(613\) images. In both phases, we used crowd-sourcing techniques with multiple volunteers labeling the dataset using the open-source tool LabelImg \footnote{https://github.com/tzutalin/labelImg}. For all the images in the dataset volunteers were provided Table \ref{tab:categories} as well as examples of correctly labeled images, incorrectly labeled images, and not applicable images. Before the labeling task, each volunteer was provided with an exercise to verify if the volunteer was able to correctly identify categories as well as identify if an annotated image is correctly labeled, incorrectly labeled, or not applicable.

The labeling process first involved two volunteers independently labeling an image from the dataset. In any of the cases where the number of bounding boxes is different, the labels for on or more of the bounding boxes are different or two volunteer annotations are sufficiently different; a third volunteer compiles the result from the two annotations to come up with a correctly labeled image. After this step, a volunteer verifies the bounding box annotations. Following this method of labeling the dataset we ensured that all images were labeled accurately and contained exhaustive annotations. As a result of this, our dataset consists of \(1029\) high-quality, majorly non-iconic, and accurately annotated images.

In Table \ref{tab:frequency-labels} we show the frequency of the categories in the Medical Personal Protective Equipment (CPPE-5) dataset. \cls{Gloves} and \cls{Mask} are the most common annotations, with a considerable portion of the bounding boxes being marked as such.

\begin{table}[ht]
    \caption{Frequency of the categories appearing in the CPPE-5 dataset calculated by the percentage of bounding boxes.}
    \centering
    \begin{tabular}{cccccc}
    \toprule
    Category & Coverall & Mask & Goggles & Face\_Shield & Gloves \\
    \midrule
    Frequency & 25.48 \% & 27.76 \% & 8.66 \% & 9.51 \% & 28.59 \% \\ 
    \bottomrule
    \end{tabular}
    \label{tab:frequency-labels}
\end{table}

\section{Dataset Statistics}
\label{Dataset Statistics}

Next, we analyze the properties of the Medical Personal Protective Equipment (CPPE-5) dataset. The Medical Personal Protective Equipment (CPPE-5) dataset contains 1029 images and 4698 object annotations consisting of 1343 glove annotations, 1304 mask annotations, 1197 coverall annotations,  447 face shield annotations, and 407 goggle annotations as shown in Table \ref{tab:annotations}. Table \ref{tab:annotations} also includes the number of images that contain at least 1 annotation belonging to a specific category.

\begin{table}
    \caption{Number of annotations in the dataset}
    \centering
    \begin{tabularx}{\textwidth}{lccc}
    \toprule
    Category     &   No. of annotations & No. of images with \(\geq\) 1 category annotation & Average annotations/image \\
    \midrule
     Coverall    &   1197 & 799 & 1.50 \\
     Mask        &   1304 & 898 & 1.45 \\
     Goggles     &    407 & 312 & 1.30 \\
     Face\_Shield &    447 & 344 & 1.30 \\
     Gloves      &   1343 & 575 & 2.34 \\
    \midrule
     Total       &   4698 & & 4.57 \\
    \bottomrule
    \end{tabularx}
    \label{tab:annotations}
\end{table}

We also compare the goals of the Medical Personal Protective Equipment (CPPE-5) dataset with other previous object detection datasets namely ImageNet \cite{deng2009imagenet}, PASCAL VOC 2012 \cite{pascal-voc-2012}, Microsoft COCO \cite{lin2014microsoft} and RMFD \cite{wang2020masked}. ImageNet's goals include capturing a large number of object categories, many of which are fine-grained. PASCAL VOC’s goals include object detection in natural images. Microsft COCO is designed for the detection of objects occurring in their natural context. Real-world Masked Face Recognition Dataset aims to detect masked faces. The Medical Personal Protective Equipment (CPPE-5) dataset is designed for subordinate object detection for Personal Protective Equipment.

\begin{figure}
    \centering
    \includegraphics[width=\textwidth]{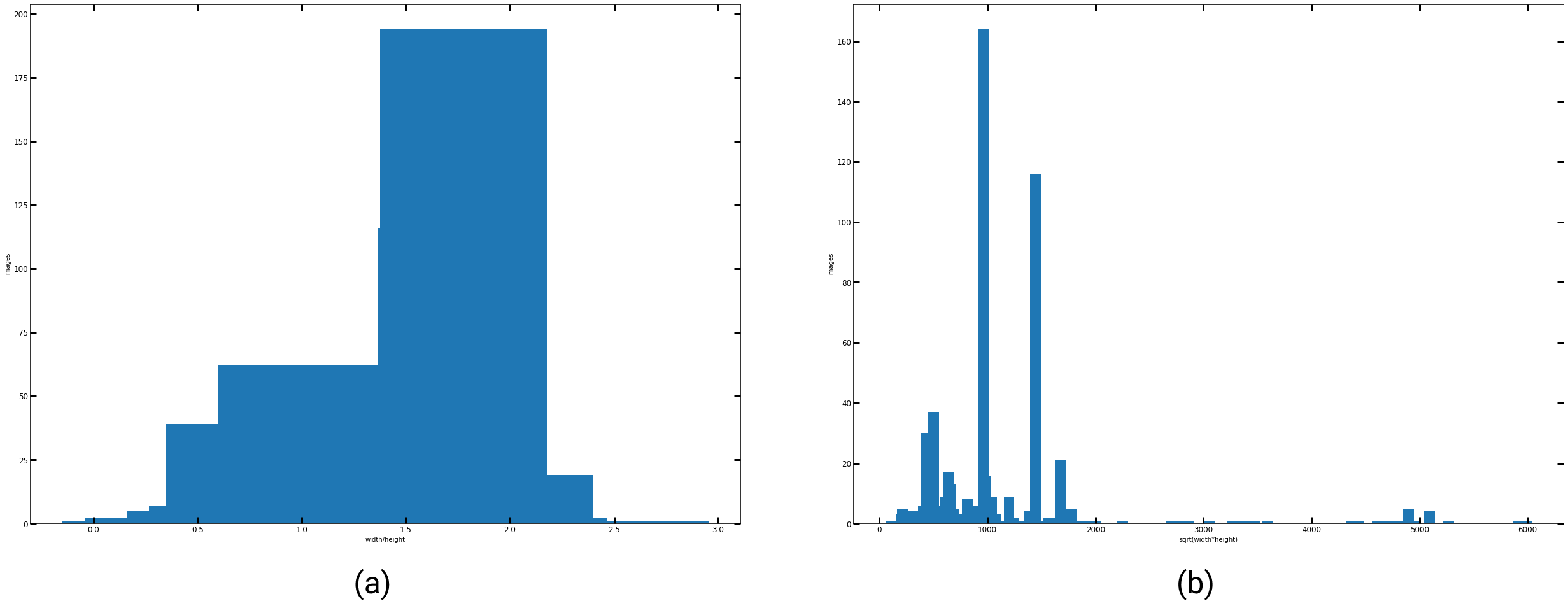}
    \caption{Image statistics, (a) distribution of aspect ratios as measured by \(\frac{width}{height}\) and (b) distribution of image sizes as measured by \(\sqrt{width \times height}\). Overall the average aspect ratio is 1.40 and the image size is 946.94 pixels.}
  \label{fig:image-stats}
\end{figure}

Next, we present statistics for images present in the dataset. On average our dataset contains 4.57 annotations per image. As shown in Fig. \ref{fig:image-stats} (a) we calculate the distribution of aspect ratios as measured by \(\frac{width}{height}\). Our dataset has an average aspect ratio of \(1.40\). We also measure the distribution of image sizes as measured by \(\sqrt{width \times height}\). Generally smaller objects are harder to recognize and require more contextual reasoning to recognize, our dataset has an average image size of 946.94 pixels.

\section{Experimental Results}
\label{Experimental Results}

In this section, we evaluate baseline and state-of-the-art object detection models trained on the Medical Personal Protective Equipment Dataset (CPPE-5) through extensive experiments. In Sec. \ref{Experimental Setup}, we detail the experimental setup used. In Sec. \ref{Baseline Models}, we describe how we chose the baseline models and share the results of the baseline models. In Sec. \ref{Evaluating State-of-the-Art models}, we present results for State-of-the-Art object detection techniques trained on the Medical Personal Protective Equipment Dataset (CPPE-5) and make inferences about the difficulty of the dataset. To foster easy reproducibility of the results we present in this section, we have open-sourced the training code, trained models as well as the training logs as TensorBoard \cite{tensorflow2015-whitepaper} dashboards in the associated code repository.

\subsection{Experimental Setup}
\label{Experimental Setup}

Our experiments are based on the open-source detection toolbox MMDetection \cite{mmdetection} and implementations from the TensorFlow Model Garden \cite{tensorflowmodelgarden2020}. The training is conducted on the \(1029\) training images and the models are tested using another set of \(100\) testing images. Depending on the throughput the models were trained either on 8 Tesla A100 GPUs or on a Cloud TPUv3 cluster.

For evaluation, we adopt the metrics from the COCO detection evaluation criteria, including the mean Average Precision (AP) across IoU thresholds ranging from \(0.50\) to \(0.95\) at different scales which are standard for object detection tasks. The inference speed FPS (Frames per second) for the detector is measured on a machine with 1 Tesla V100 GPU.

\subsection{Baseline Models}
\label{Baseline Models}

\begin{table}
    \caption{Baseline models trained on the CPPE-5 dataset.}
    \centering
    \begin{tabular}{c|c|ccccc|c|c}
        \toprule
        Method & \(AP^{box}\) & \(AP^{box}_{50}\) & \(AP^{box}_{75}\) & \(AP^{box}_{S}\) & \(AP^{box}_{M}\) & \(AP^{box}_{L}\) & \#Params & Epochs \\
        \hline
        SSD \cite{10.1007/978-3-319-46448-0_2} & 29.50 & 57.0 & 24.9 & 32.1 & 23.1 & 34.6 & 64.34 M & 160 \\
        YOLO \cite{redmon2018yolov3} & 38.5 & 79.4 & 35.3 & 23.1 & 28.4 & 49.0 & 61.55 M & 273 \\
        Faster RCNN \cite{7485869} & 44.0 & 73.8 & 47.8 & 30.0 & 34.7 & 52.5 & 60.14 M & 24 \\
        \bottomrule
    \end{tabular}
    \label{tab:baseline}
\end{table}

A significant gain was obtained in object detection with the introduction of Regions with CNN features (R-CNN). DNNs, or the most representative CNNs, act in a quite different way from traditional approaches. They have deeper architectures with the capacity to learn more complex features than shallow ones. R-CNN \cite{girshick2014rich} brought the advances in image classification using deep learning to object detection using a two-stage approach: classify object proposal boxes into any of the classes of interest.

Since the proposal of R-CNN, a lot of improved models have been suggested, including Fast R-CNN which jointly optimizes classification and bounding box regression tasks, and Faster R-CNN which takes an additional sub-network to generate region proposals. Faster R-CNN stills provide very competitive results today in terms of accuracy. More recently, single-shot detectors were presented to bypass the computational bottleneck of object proposals by regressing object locations directly from a predefined set of anchor boxes (e.g. SSD \cite{10.1007/978-3-319-46448-0_2} and YOLO \cite{redmon2018yolov3}). This typically results in simpler models that are easier to train end-to-end \cite{zhao2019object, kuznetsova2020}. All of them bring different degrees of detection performance improvements over the primary R-CNN and make real-time and accurate object detection become more achievable.

We carefully choose Faster-RCNN \cite{7485869}, YOLOv3 \cite{redmon2018yolov3} and SSD \cite{10.1007/978-3-319-46448-0_2} as our baseline testing algorithms for their excellent performance on general object detection. In Table \ref{tab:results} we present the results for these three baseline models.

\begin{description}[style=unboxed,leftmargin=0cm]
\item[Faster RCNN] was trained with a ResNet 101 backbone. Only random flip data augmentations  were applied to the image. We use the SGD optimizer with a momentum of \(0.9\) and a weight decay of \(0.0001\) and no gradient clipping. We use a step learning rate scheduler with an initial learning rate of \(0.02\) with a linear warm-up for \(500\) iterations. We use sigmoid cross entropy loss as the classifier loss and L1 loss as the bounding box loss. This baseline model was trained for 24 epochs.
\item[YOLO] was trained with a DarkNet 53 backbone The data augmentation pipeline uses random flip, photometric distortion, and a random crop on the image and bounding boxes such that the cropped patches have minimum IoU requirement with the original image and bounding boxes. We use the SGD optimizer with a momentum of \(0.9\) and a weight decay of \(0.0005\) and apply gradient clipping using the \(L^2\) norm. We use a step learning rate scheduled with an initial learning rate of \(0.001\) with a linear warmup for \(2000\) iterations. We use sigmoid cross entropy loss as the classifier loss, confidence loss, and the xy-coordinate loss, and MSE loss for wh-coordinate loss. The xy-coordinate loss and wh-coordinate loss use a weight of \(2\). This baseline model is trained for 273 epochs.
\item[SSD] was trained with a MobileNet V1 backbone . Only random flip data augmentations  were applied to the image. We use the momentum optimizer with a momentum of \(0.9\). We use a cosine decay learning rate schedule with an initial learning rate of \(0.04\) and warmup for \(2000\) iterations with the learning rate \(\frac{4}{300}\). We use weighted smoothed \(L^{1}\) as the localization loss and weighted sigmoid focal as the classification loss with \(\alpha=0.25\) and \(\gamma=2.0\). This baseline model is trained for 160 epochs.
\end{description}

\subsection{Evaluating State-of-the-Art models}
\label{Evaluating State-of-the-Art models}

\begin{table}
    \caption{Top performing models based on the standard metric, box AP, trained on the CPPE-5 dataset.}
    \centering
    \begin{tabularx}{\textwidth}{c|c|ccccc|c|c}
        \toprule
        Method & \(AP^{box}\) & \(AP^{box}_{50}\) & \(AP^{box}_{75}\) & \(AP^{box}_{S}\) & \(AP^{box}_{M}\) & \(AP^{box}_{L}\) & \#Params & Epochs \\
        \hline
        RepPoints \cite{yang2019reppoints} & 43.0 & 75.9 & 40.1 & 27.3 & 36.7 & 48.0 & 36.6 M & 24 \\
        Sparse RCNN \cite{sun2021sparse} & 44.0 & 69.6 & 44.6 & 30.0 & 30.6 & 54.7 & 124.99 M & 36 \\
        FCOS \cite{tian2019fcos} & 44.4 & 79.5 & 45.9 & 36.7 & 39.2 & 51.7 & 50.8 M & 24 \\
        Grid RCNN \cite{8953602, lu2019grid} & 47.5 & 77.9 & 50.6 & 43.4 & 37.2 & 54.4 & 121.98 M & 25 \\
        Deformable DETR \cite{zhu2020deformable} & 48.0 & 76.9 & 52.8 & 36.4 & 35.2 & 53.9 & 40.5 M & 50 \\
        FSAF \cite{zhu2019feature} & 49.2 & 84.7 & 48.2 & 45.3 & 39.6 & 56.7 & 93.75 M & 12 \\
        Localization Distillation \cite{zheng2021localization} & 50.9 & 76.5 & 58.8 & 45.8 & 43.0 & 59.4 & 32.05 M & 12 \\
        VarifocalNet \cite{Zhang_2021_CVPR} & 51.0 & 82.6 & 56.7 & 39.0 & 42.1 & 58.8 & 53.54 M & 24 \\
        RegNet \cite{radosavovic2020designing} & 51.3 & 85.3 & 51.8 & 35.7 & 41.1 & 60.5 & 31.5 M & 24 \\
        Double Heads \cite{wu2020rethinking} & 52.0 & 87.3 & 55.2 & 38.6 & 41.0 & 60.8 & 148.7 M & 12 \\
        DCN \cite{dai2017deformable, zhu2019deformable} & 51.6 & 87.1 & 55.9 & 36.3 & 41.4 & 61.3 & 148.71 M & 12 \\
        Empirical Attention \cite{Zhu_2019_ICCV} & 52.5 & 86.5 & 54.1 & 38.7 & 43.4 & 61.0 & 47.63 M & 12 \\
        \textbf{TridentNet} \cite{li2019scale} & \textbf{52.9} & \textbf{85.1} & \textbf{58.3} & \textbf{42.6} & \textbf{41.3} & \textbf{62.6} & 32.8 M & 36 \\
        \bottomrule
    \end{tabularx}
    \label{tab:results}
\end{table}

We also present results from training some state-of-the-art object detection models on Medical Personal Protective Equipment Dataset (CPPE-5) in Table \ref{tab:results} using the same evaluation procedure as mentioned earlier. Comparing these results with that of some other widely used object detection datasets like OpenImages, Microsoft COCO, and Pascal VOC \footnote{https://paperswithcode.com/sota}, we conclude that Medical Personal Protective Equipment Dataset (CPPE-5) does include more difficult (non-iconic) images of objects. We include more details on how each of these models was trained in the associated code repository.

\section{Conclusion}
\label{Conclusion}

This paper presented a new object detection dataset, the Medical Personal Protective Equipment Dataset (CPPE-5) which is the first dataset focusing on the subordinate category of medical Personal Protective Items and would have wide practical uses. We conducted a detailed analysis of the dataset and compared it to other popular broad-category datasets and datasets focusing on personal protective equipment. We found that there is currently no publicly available dataset for studying subordinate categorization of medical personal protective equipment. Overall, our CPPE-5 dataset fills a significant gap in the availability of datasets for the study of subordinate categorization of medical personal protective equipment. We annotate a huge number of well-distributed oriented objects with oriented bounding boxes with emphasis placed on finding non-iconic images of objects in natural environments and varied viewpoints. We assume this dataset is challenging but very similar to real-world scenarios, making this an appropriate dataset for practical applications. We explained how the data was collected and annotated and presented dataset statistics indicating that the images often contain multiple bounding boxes per image. We further also evaluated multiple modern state-of-the-art and baseline object detection models trained on our dataset, establishing a benchmark for subordinate categorization for medical Personal Protective Equipment images. Many object detection algorithms benefit from additional annotations, such as the amount an instance is occluded or the location of key points on the object which we believe are promising directions for future annotations. Detecting medical Personal Protective Equipments is a task of great practical importance, we believe CPPE-5 will not only promote the development of object detection algorithms for this purpose but also pose interesting algorithmic questions to general object detection in computer vision.

\section*{Data Availability}
\label{Data Availability}

The datasets generated during and/or analyzed during the current study are available in the CPPE-5 repository, at \url{https://git.io/cppe5-dataset}.

\section*{Acknowledgements}
\label{Acknowledgements}

The authors would like to thank Google for supporting this work by providing Google Cloud credits. The authors would also like to thank Google TPU Research Cloud (TRC) program \footnote{https://sites.research.google/trc} for providing access to TPUs. The authors are also grateful to Omkar Agrawal for his help with verifying the difficult annotations.

\section*{Competing Interests}
The authors declare no competing interests.

\bibliographystyle{plain}
\bibliography{references.bib}

\newpage

\section*{Appendix Overview}
\label{Appendix Overview}

In the appendix, we provide the implementation details for our experiments in Appendix \ref{Implementation Details}, some more sample images in Appendix \ref{appendix:sample-images} and compare the model complexities for each of the models we presented in the Sect. \ref{Experimental Results} with their performance on the CPPE-5 dataset in Appendix \ref{appendix:Comparing model complexities}.

\appendix

\section{Implementation Details}
\label{Implementation Details}

In this section, we explain the implementation details of the experiments we perform and the models we train.

\paragraph{Sampling.} There is a slight class imbalance in the dataset for some of the classes, meaning that not all classes have a similar number of images. For this reason, we follow a stratified sampling strategy during data loading.

\paragraph{Code. } Our code is in PyTorch 1.10 \cite{paszke2019pytorch}. We use a number of open-source packages to develop our training workflows. Most of our experiments and models were trained with mmdetection \cite{mmdetection} and we also used timm \citep{rw2019timm} for some of the experiments. We also utilized TensorFlow \cite{45381}, TensorFlow Lite \footnote{\url{https://github.com/tensorflow/tensorflow/tree/master/tensorflow/lite}}, and TensorFlow.js \footnote{\url{https://github.com/tensorflow/tfjs}} for creating edge deployment ready models for the mobile and browser. Furthermore, we also used Tensorboard \cite{tensorflow2015-whitepaper} while training the model. Our hardware setup for the experiments included either eight NVIDIA Tesla A100 GPUs or a TPUv3 cluster. We utilized mixed-precision training with PyTorch's native AMP (through \texttt{torch.cuda.amp}) for mixed-precision training and a distributed training setup (through \texttt{torch.distributed.launch}) which allowed us to obtain significant boosts in the overall model training time.

\paragraph{Hyperparameters.} Due to the extent of our experiments, we redirect the reader to our GitHub repository to find the hyperparameters and configurations for each of the experiments in this paper.

\section{Sample Images}
\label{appendix:sample-images}

\begin{figure}
    \centering
    \includegraphics[height=0.9\textheight]{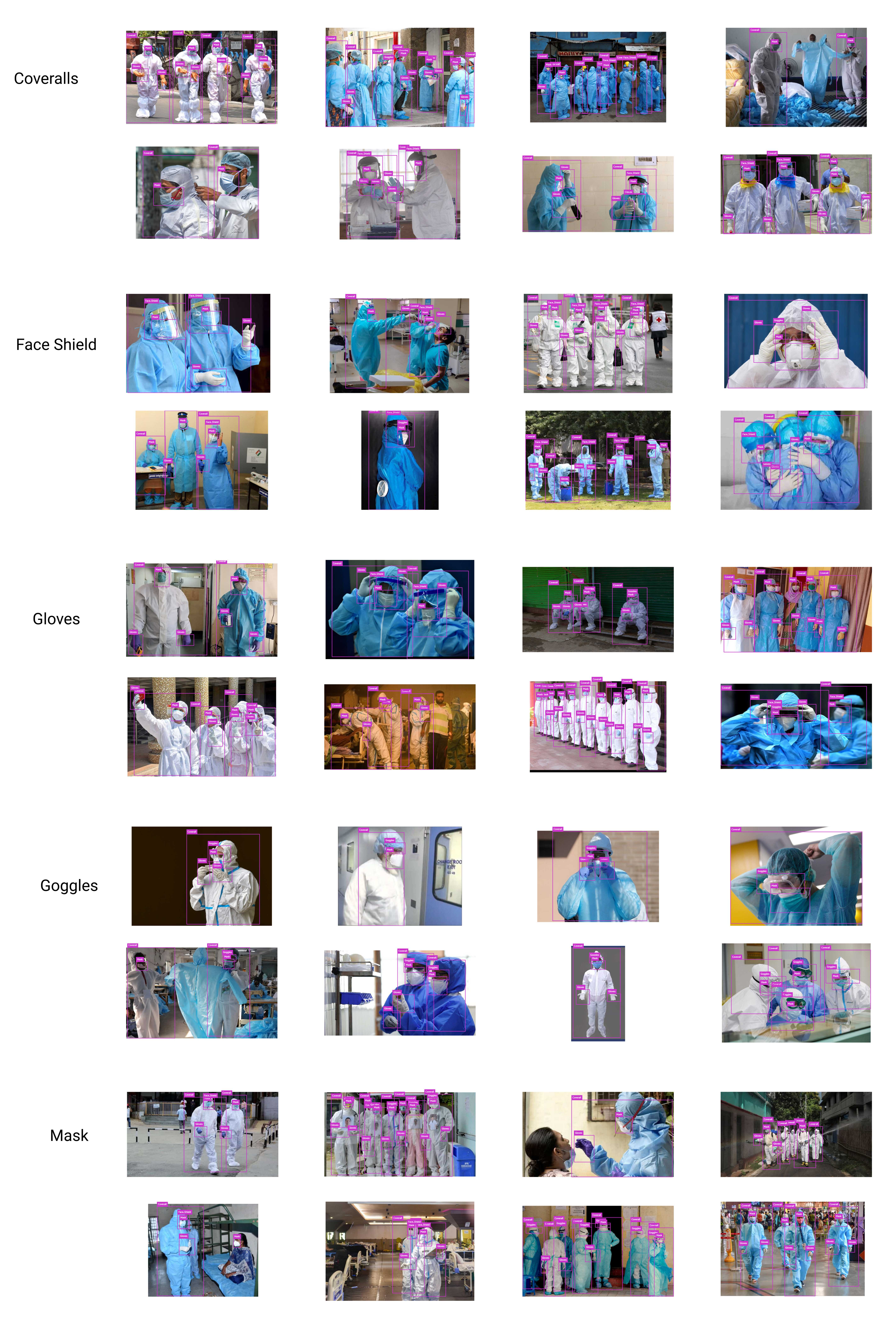}
    \caption{Samples of annotated images for each category in the CPPE-5 dataset.}
  \label{fig:sample-images}
\end{figure}

In Fig. \ref{fig:sample-images} we show 8 sample images from each of the categories in the dataset with the object annotations superimposed on the images. It is noteworthy to know that some of the images may not have the original image sizes since the class names were superimposed on the image and we did not want the class names to be cut off. These visualizations were generated with FiftyOne \cite{moore2020fiftyone}.

\section{Comparing model complexities}
\label{appendix:Comparing model complexities}

\begin{figure}
    \centering
    \includegraphics[width=0.8\textwidth]{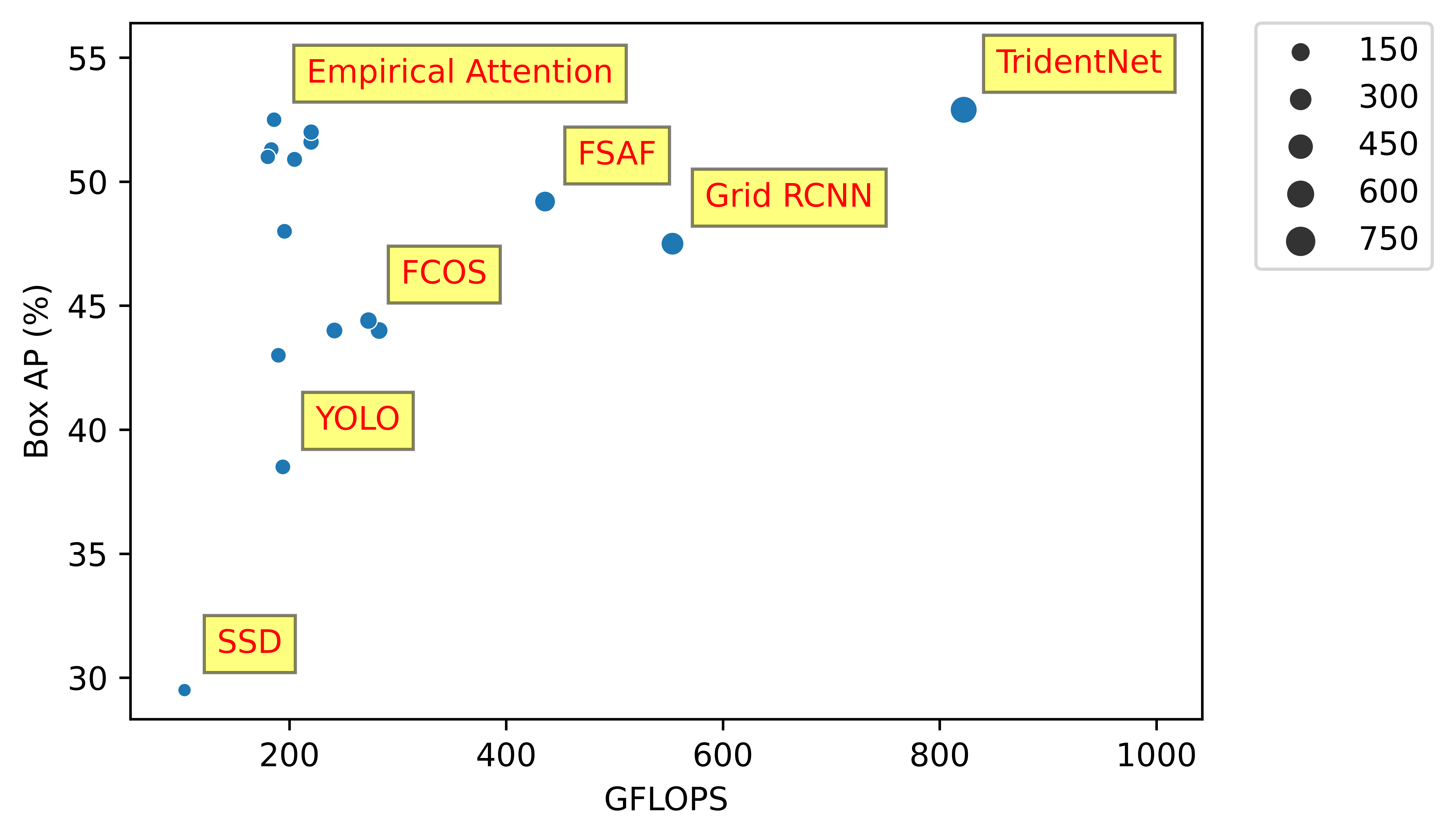}
    \caption{Comparison of model performance and model complexity in terms of FLOPs (in billions).}
  \label{fig:flops}
\end{figure}

\begin{figure}
    \centering
    \includegraphics[width=0.8\textwidth]{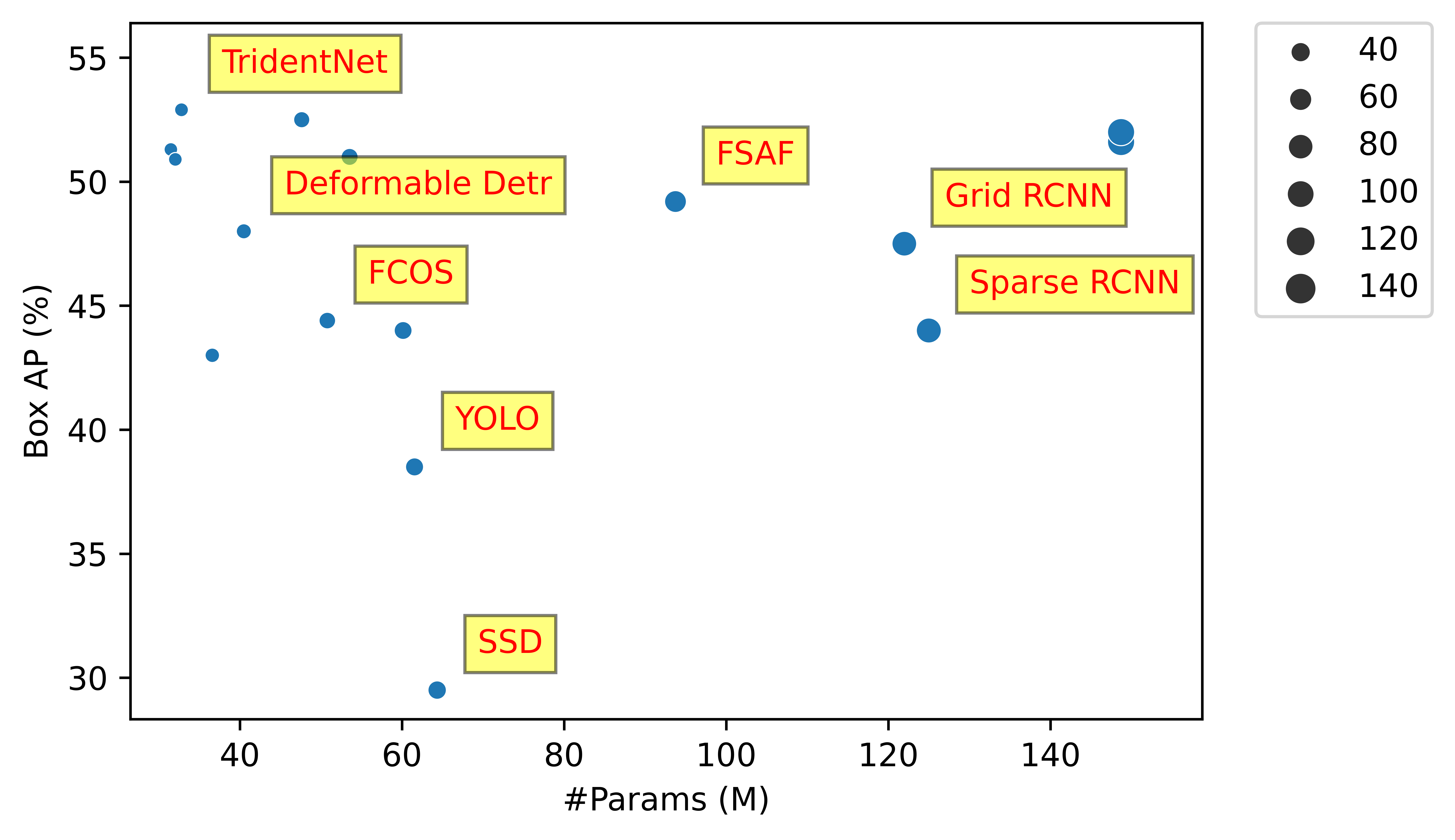}
    \caption{Comparison of model performance and model complexity in terms of number of parameters (in millions).}
    \label{fig:params}
\end{figure}

In Table \ref{tab:model-complexity} and we compare model complexities and their performance on the CPPE - 5 (Medical Personal Protective Equipment) dataset. We measure model complexity in terms of the number of parameters of the model and FLOPs required to run a single instance of the model. In Fig. \ref{fig:flops} and Fig. \ref{fig:params} we show a visual representation of comparing the model complexities.

\begin{table}[h]
    \caption{Comparison between model complexity, in terms of number of parameters (in millions), FLOPs (in billions), and frames per second on a Tesla V100 GPU, and \(AP^{box}\).}
    \centering
    \begin{tabularx}{\textwidth}{c|YYYY}
        \toprule
        Method & \(AP^{box}\) & \#Params & FLOPs & FPS \\
        \hline
        SSD & 29.5 & 64.34 M & 103.216 G & 25.6 \\
        YOLO & 38.5 & 61.55 M & 193.93 G & 48.1 \\
        RepPoints & 43.0 & 36.6 M & 189.83 G & 18.8 \\
        Faster RCNN & 44.0 & 60.14 M & 282.75 G & 15.6 \\
        Sparse RCNN & 44.0 & 124.99 M & 241.53 G & 21.7 \\
        FCOS & 44.4 & 50.8 M & 272.93 G & 9.7 \\
        Grid RCNN & 47.5 & 121.98 M & 553.44 G & 7.7 \\
        Deformable DETR & 48.0 & 40.5 M & 195.47 G & 18.8 \\
        FSAF & 49.2 & 93.75 M & 435.88 G & 5.6 \\
        Localization Distillation & 50.9 & 32.05 M & 204.71 G & 19.5 \\
        VarifocalNet & 51.0 & 53.54 M & 180.05 G & 4.8 \\
        RegNet & 51.3 & 31.5 M & 183.29 G & 18.2 \\
        Double Heads & 52.0 & 148.7 M & 220.05 G & 9.5 \\
        DCN & 51.6 & 148.71 M & 219.97 G & 16,6 \\
        Empirical Attention & 52.5 & 47.63 M & 185.83 G & 12.7 \\
        TridentNet & 52.9 & 32.8 M & 822.13 G & 4.2 \\
        \bottomrule
    \end{tabularx}
    \label{tab:model-complexity}
\end{table}

\end{document}